# A comprehensive review and evaluation on text predictive and entertainment systems

Hozan K. Hamarashid[1], Soran A. Saeed[2], Tarik A. Rashid[3]
[1] Computer Science Institute, Sulaimani Polytechnic University, Sulaimani, Iraq
[2] Sulaimani Polytechnic University, Sulaimani, Iraq
[3] Computer Science and Engineering, University of Kurdistan Hewler, Erbil, Iraq

**Abstract**

One of the most important ways to experience communication and interact with the systems is by handling the prediction of the most likely words to happen after typing letters or words. It is helpful for people with disabilities due to disabling people who could type or enter texts at a limited slow speed. Also, it is beneficial for people with dyslexia and those people who are not well with spells of words. Though, an input technology, for instance, the next word suggestion facilitates the typing process in smartphones as an example. This means that when a user types a word, then the system suggests the next words to be chosen in which the necessary word by the user. Besides, it can be used in entertainment as a gam, for example, to determine a target word and reach it or tackle it within 10 attempts of prediction. Generally, the systems depend on a text corpus, which was provided in the system to conduct the prediction. Writing every single word is time-consuming, therefore, it is vitally important to decrease time consumption by reducing efforts to input texts in the systems by offering most probable words for the user to select, this could be done via next word prediction systems. There are several techniques can be found in literature, which is utilized to conduct a variety of next word prediction systems by using different approaches. In this paper, a survey of miscellaneous techniques towards the next word prediction systems will be addressed. Besides, the evaluation of the prediction systems will be discussed. Then, a modal technique will be determined to be utilized for the next word prediction system from the perspective of easiness of implementation and obtaining a good result.

1. **Introduction**

In the field of machine learning, various technologies have been developed in today's world, for instance, next word prediction is an assistance system for typing. In other words, it reduces efforts and lessens time-consuming during typing. Besides that, some people have difficulty with slow typing speed because of their disabilities, dyslexia, or memorizing spells of words, thus, the next word suggestion has been developed to help the mentioned people. Also, it can be used in entertainment computing, such as computer games, which is helpful for people to relax and enjoy after tiredness.

Word prediction system is about suggestion next words or most probable words to happen based on what user types or in a sequence of words. This depends on a text corpus (Carlberger, 1997). These types of systems suggest most likely to happen either letters or words while the letters or words are entered by the user (Matthew, 1996). When a user types a character or a word, the system suggests the most common or most frequent letter or word after the typed character or word then the next words would appear after the word has been selected by the user. This process will continue until the text is accomplished (Ghayoomi and Momtazi, 2009).

Word prediction can be utilized in entertainment computing. Entertainment computing includes the fields of computer games, deliver mobile contentment as entertainment, media interaction, robotics entertainment, entertainment for psychology and sociology, and entertainment for augmented or virtual reality (Nakatsu and Hoshino, 2013). Nowadays, the essential area of computing is entertainment computing (Wong, 2008). Currently, it can be seen that there is a focus on entertainment and its value perceived for individuals (Rauterberg, 2009). In addition, next word prediction can be utilized for entertainment and games depending on artificial intelligence to design the game (Treanor et al, 2015). Entertainment computing and game design methodology utilize artificial intelligence and algorithms for making games and their foreground as well as using them in the background of the game (Andersen et al, 2018). Making entertainment games by using predictive text, which utilize AI agents and Natural Language Processing NLP (Yannakakis and Hallam, 2006). The player is capable to interact with the game by writing or posting a message or entering a target word into the system. Therefore, to make progress, the player requested to compose the estimated text into the game. In this step, the player will be expedited by the system by providing a predictive text to the interface of the game. This helps the player to look at what has been composed recently. In addition, in the game real-time feedback was provided to select the words that most likely to utilize in the next step (Campton et al, 2015).

The main difficulty is about choosing the best technique for the next word prediction. However, there were several techniques to conduct the next word suggestion but each of one them has difficulties in the aspect of applying the model, configuring data sets, etc. N-grams as an example, which number of $N$ is better to be chosen to conduct a better word prediction. On the other hand, the text corpus has a vitally important role, the bigger size of the text corpus will lead to have adequate data and slightly decrease performance. Though this paper helps to illustrate different techniques with advantages and drawbacks also it helps to determine a better approach to be elected for conducting the next word prediction system.

As a consequence, the next word prediction could be either word completion or word suggestion. So, word completion is a system that suggests words when a character has been entered. On the other hand, word suggestion is a system that suggests the most likely to happen words to the user when a word has been entered or selected but it depends on earlier typed words instead of the basis of the character.

The main aim of word prediction systems is to improve KSR (keystroke saving rate). KSR is the number of saved keystrokes by the user during using the word prediction system. Better performance will be obtained when the number of KSR is high. Namely, increasing the number of KSR will increase performance (Khan et al., 2009). Consequently, less time and effort are needed to obtain a completed text. Also, the next word prediction system is helpful for people with disabilities, dyslexia or those people that often make mistakes in the spell of words due to the system requires few characters to type rather than the whole word or words.

Different types of next word prediction systems were developed with various techniques. So, in this paper, different techniques for next word prediction systems are addressed, also evaluation of next word prediction systems are discussed. Besides, the next word prediction approaches and its taxonomy are addressed then entertainment computing with text predictive systems is discussed.

Also, the paper is structured as follows: Section 2, presents the Taxonomy of various techniques that are developed to predict the next words. In section 3, an in-depth analysis of the mentioned techniques in section 2 or a literature survey, is illustrated and the consequences of each technique are addressed. Section 4 discusses entertainment computing and text prediction. Section 5 illustrates the analysis of the next word prediction techniques. Open research problems and challenges are discussed in section 6. Finally, concluding points and future works of this paper are outlined in section 7.

## 2. Taxonomy of next word prediction techniques

In the recent development of next word prediction with entertainment computing, the following categories are noticed: statistical methods and artificial intelligence. Mostly, these categories go under expansive types of technical analysis. Sometimes, these categories are combined to predict the next word also they can be utilized to make games in entertainment computing. For instance, these broader techniques of technical analysis can be combined with deep learning algorithms. These approaches have achieved good results and become popular in the field of next word prediction for entertainment computing, such as games and social predictive texts in the current and past. Popular methods are shown in Figure1:

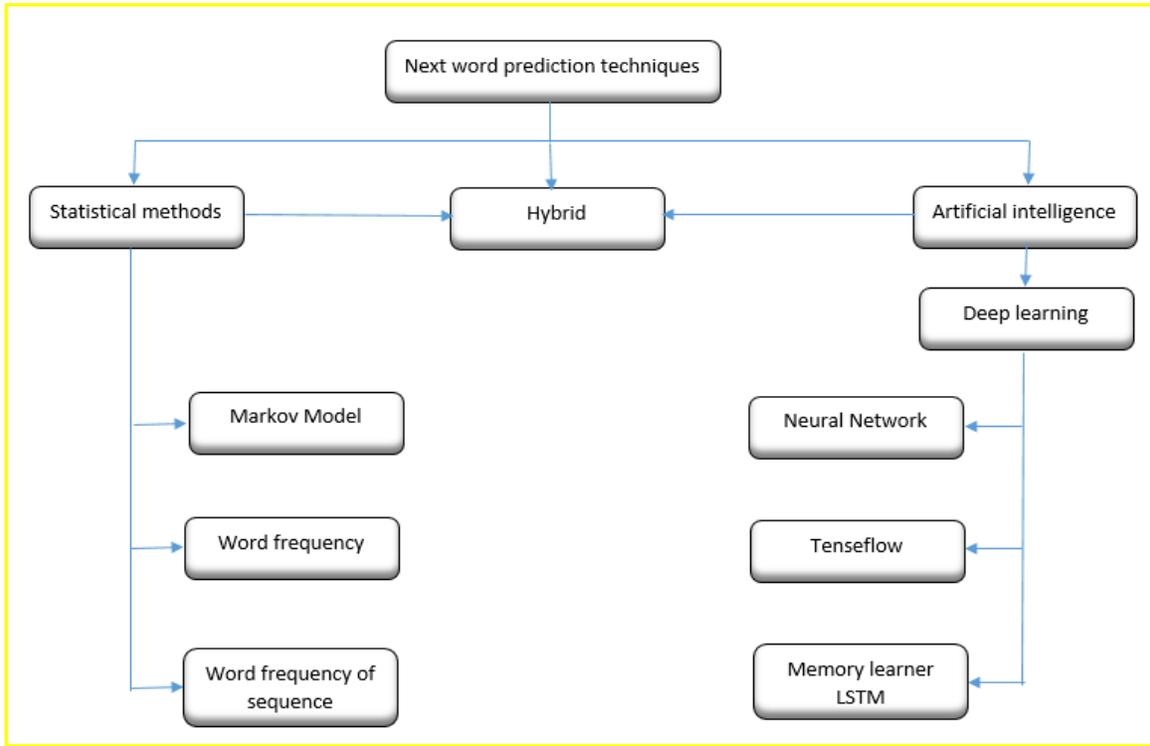

**Figure1: Taxonomy of next word prediction systems**

Previously, before the deep learning techniques were advanced, statistical methods merely were utilized to predict the next words or to build next word prediction systems for entertainment computing. Markov Chain is the first method used to predict the next word in natural language processing NLP and entertainment computing. The goal of NLP is to learn and analyze the difficulties of automated generation and comprehending of languages of human beings (Panzner and Cimiano, 2016). Markov Model is the origin of the next word prediction systems, which is used in the field of entertainment computing, such as games. In this model, the next word suggestion based on the probability of a word in the text corpus (Bari, 2014). Thus, to choose the next optimum word, the Markov model decides which word should be anticipated. However, the Markov model does not have a memory to decide for a long-range for instance in a long sentence (Chakraborty and Roy, 2012). It has been used based on the effectiveness and its easiness of implementation. Word frequency approach which depends on a text corpus, is another method that utilized for the next word prediction. In this method the frequencies of words are saved as a list, so a small number of words will be presented to, due to the words with frequencies are listed from high to low (Neubig, 2016). This is called the Unigram model in another word it is well known as it does not utilize history (Neubig, 2016).

Word sequence frequencies, is another method that is used to predict the next word to see which word is coming next, namely to predict the future with social media as an example (Sitaram and Huberman, 2010). In this method, the previous words are essential, unlike the unigram model, which is not depending on previous words or history words. In another word, the Bigram model approximates the frequency or probability of a word given the entire history words by utilizing merely the conditional probability of one earlier word (Kapadia, 2019). We can say that it is an extension of the Unigram model. However, if the last two previous words are utilized to predict the next word then it is known as the Trigram model (Gendron, 2015). Also, to anticipate the next word if extra words are used then it is called as N-gram model, which is very useful and powerful in certain points such as easiness of implementation and it does not depend on grammar rules. Therefore, in its significance, a statistical model for language purposes, is the model class in which the probabilities for the sequence of words will be assigned. The n-gram model as an example is a very common and well-known method. An n-gram model, predicts the most likely words to happen to a sequence when a sequence of *N-1* is given (Chelba et al., 2017). This model is very useful in NLP and building various applications such as text prediction, machine translation, and speech recognition, etc.

Artificial intelligence is huge, commonly it is about the development of computer systems and applications to accomplish tasks. Generally, it requires human intelligence, for example, speech recognition, language translation, and decision- making, etc. Besides, it can be said that AI is the competence of computer to conduct tasks. Besides, it has a big role in entertainment computing, such as mobile entertainment and games, for example, chess, poker. In addition, it includes technologies, for instance, search techniques, machine-learning games, and human computer interaction. Normally, it is correlated with intelligent beings (Singh, 2014). Namely, it is repeatedly practiced with the undertaking works and developing systems enriched with smart distinctive processes of humans as an example the capability of reasoning, finding meanings, and learn from earlier knowledge, etc. (Wang, 2008). Machine learning ML which is inside the area of AI is about studying algorithms and statistical methods in which a computer system utilizes to execute a particular task without utilizing certain instructions, instead of unambiguous instructions, patterns and inference will be depended on. Deep learning is a portion of machine learning approaches relying on artificial intelligence (Dargan et al, 2019).

Neural Network, organically or artificially, refers to the systems of neurons. Namely, an artificial neural network consists of artificial neurons. In this sense, an artificial neural network was developed to solve artificial intelligent difficulties (Mohanapriya, 2017; Jabar et al., 2018). The most common methods in the neural network are recurrent neural network RNN and convolutional neural network CNN. RNN is a type of artificial neural network ANN in which the output or results from earlier phases are fed into the systems as an input to the current step. So to predict the next word, the earlier words are needed to be remembered. As a consequence, the development of RNN conducted to solve this type of difficulty by utilizing the hidden layer in RNN and the most essential part is the hidden state. Generally, it consists of input data, hidden layer, and output. RNN can be used in time series data, text data, and audio data (Arnold et al, 2016).

Convolutional neural network CNN is a class of deep neural network, CNN is multiple perceptron due to each neuron in a layer is connected to all neurons in the next layer that is the reason of calling CNN as a fully connected network it can be used in different fields such as image processing and video analyzing (Arnold et al, 2016).

Long short term memory LSTM is a class of RNN that can learn sequencing orders in sequencing prediction difficulties. LSTM is very complicated in the way of having many parameters. Also, computationally is time-consuming. LSTM can be beneficial in the field of machine translation, speech recognition, etc. (Chen et al., 2019).

The Hybrid methods implement a mixture of multiple of various methods to enhance performance or increase accuracy, for instance, a hybrid of statistical methods and deep learning methods. Next, the most common methods which are used in the next word prediction will be explained in detail.

## 3. Literature Survey

Various techniques were developed for the next word prediction with entertainment computing. These techniques are utilized in modeling natural language processing in entertainment. Depending on the presented taxonomy in Figure 1, this survey paper demonstrates a literature survey on some of the most common approaches that have been implemented for the next word prediction for entertainment computing. Following, are the illustration of these techniques:

### 3.1 Statistical methods

In statistical approaches, to place words in the prediction list depends on the probability which will come into sight in the text corpus (Mathew, 1996). Thus, this is called in another name which is the Probabilistic approach. Next word prediction for entertainment gaming in the statistical method depends on the Markov Chain assumption. In this model, the last word $N-1$ only in the previous words or history will affect the next word (Ghayoomi and Momtazi, 2009). For that reason, this approach is also known as the N-gram model. The following are mostly utilized methods in statistical approaches:

**3.1.1 Markov chain model:** in the Markov chain model, the next word prediction in a game is depending on the probability of the text, in other words, it depends on the probability of the word in the text (Jagadeesh, 2018). Thus, it is known as a probabilistic model or statistical model. On the other hand, the statistical model depends on the Markov model. In the model, the next word merely affected by the last word $N-1$ of history (Mahar and Memon, 2011). In other words, the Markov chain can be simulated in a series of events. Every event in the series arises from a set of consequences that depends on one another. Every consequence dictates the result which is the most likely to happen next. So, in the Markov chain model to suggest the next word, all the series information is required which is included in the most current series information. Namely, by knowing all the series of previous information to predict the next word in the entertainment gaming, does not mean that it provides better results due to it depends only on the last consequence. So, the Markov chain model is used to test a consequence of words, these words are accompanying to one another with a fixed probability. Contradicts, for the dependent sequence of words, the restricted number of consequences could be raised. Though, the calculation of conditional probabilistic which is related to every consequence to one another can be conducted. Mostly, it takes the shape of calculating how many times a certain consequence pursues another one in an observed series. Consequently, to manipulate a simulation on a text corpus, every word that is utilized has to be counted. Later, each word in the text will be stored which is used next, this means the distribution of the word in the text corpus conditional on the previous word. The formula of the Markov model is written as shown below:

$$P(W|W_{t-1}, W, \ldots W_n) \approx P(W_1|W_{t-1}) \tag{1}$$

The probability that a word will be predicted given n past words nearly equals to the probability that a word will come to show given merely the last past word.

$$W_1, W_2, \ldots, W_n = \sum_n^2 P(W_t|W_{t-1}) \tag{2}$$

The second part of the Markov Chain model contains the last two words that occur in a sequence.

To reveal how it does work in the next word prediction is an example is taken into consideration as follows:

1) I study IT.
2) I study AI.
3) I love statistics.

From the above example, the unique words in the sentences are "I", "study", "love", and "IT", "AI", "statistics" might form a various state. The distribution of the probability is conclusive of the probability of transition from one word to another word. In this instance, the first word will be "I" due to the probability of occurrence of "I" is 100%. For the next or second the words "study" or "love" should be chosen. So the distribution of the probability is about the probability of the next word will be "study" or "love" given the earlier word which is "I". It can be seen that the word "study" occurs 2 times out of 3 sentences preceding by the word "I" but the word "Love" occurs merely once. Thus, the approximate probability of occurring "study" after the word "me" will be (2/3) which is 67%. In contrast, (1/3) or 33% for the word "love". Likewise, the same probability of occurrence will be for the words "IT" and "AI" to happen after "study" and the word "love" preceded by "statistics" consistently in the given example.

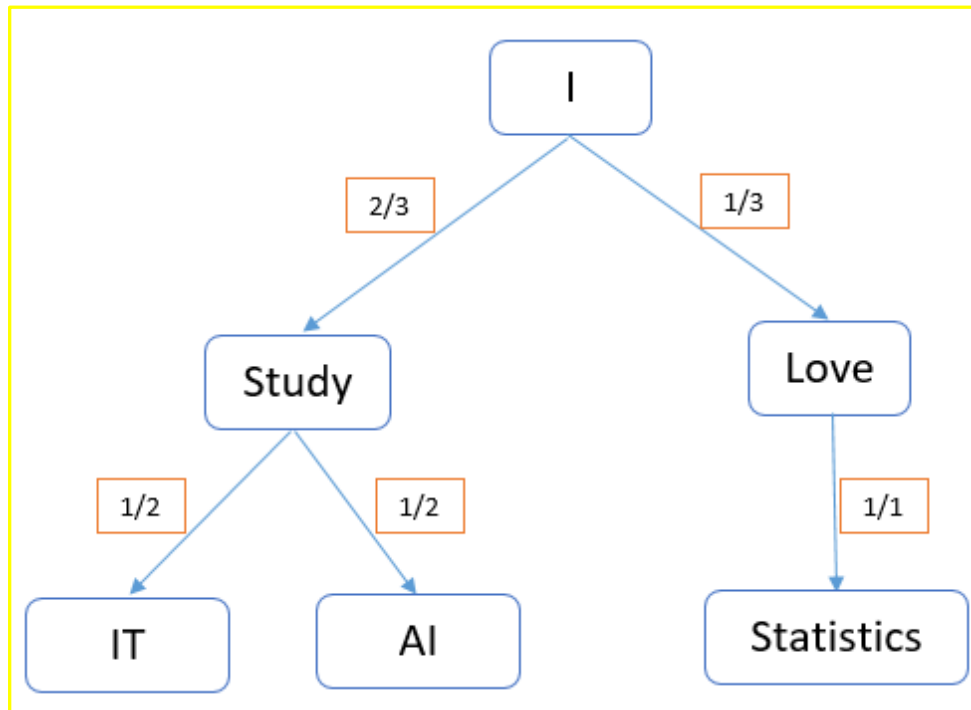

**Figure 2: the transition from one word to another in the given example**

Conditional probability of the above demonstrated example

P(study | I) = 0.67

P(love | I) = 0.33

P(AI | study) = P(IT | study) = 0.50

P(statistics | love) = 1

**3.1.2 Frequencies of words:** in this approach, the entire corpus is sorted through their frequencies order and it displays the top few words list, as next word prediction in the entertainment game to a user. This is called the unigram model (Ghayoomi and Momtazi, 2009). Natural language processing for entertainment computing depends on text corpus consistently due to the computation of word frequencies is based on the text corpus (Mahar and Memon, 2011). A dictionary of the most common words will be produced and utilized. Being a part of this step a huge text corpus will be provided.

Then, the frequency of the commonly used words or dictionary words can be calculated, which is happening in the text corpus. This means that the unigram frequency model is calculated based on the text corpus. It is possible if words exist in the text corpus but absent in the dictionary so it could be to ignore these words. Contradicts, these words that exist in the commonly used words list and missed in the text corpus cannot be ignored. The unigram language model can behave as a mixture of assorted ones state. It splits the frequency or the probability of various words in a text corpus, for example, a measure of the frequency of a word W is the number of occurrence of that word in a text corpus $C$.

$$W\ freq = \sum\ W\ where\ W\ exist\ in\ C \qquad (3)$$

This means the frequency of word $W$ = summation of occurrence of $W$ in $C$. In another way we can say to count $W$ occurrence in $C$, normally a function can be created to give $N$ when $W$ appears $N$ times in $C$. So, we can have two parameters W and $C$:

$f(C, W) = N$ when $W$ appears in $C$ $N$ times.

For the probability of Unigram model we can say as an example:

$$P(W_1\ W_2\ W_3) = P(W_1)P(W_2|W_1)P(W_3|W_1W_2)\ TO\ P(W_1\ W_2\ W_3) = P(W_1)P(W_2)P(W_3) \qquad (4)$$

In the Unigram model, every word probability relies on merely its probability in the text corpus. So, the summation of the automation probability itself over the whole of the vocabulary text corpus of the Unigram model is equal to 1. It can be represented as follows:

$$\sum_{Word\ W\ in\ corpus} P(Word\ W) = 1 \qquad (5)$$

The probability of demand to a particular word is computed like below:

$$P(demanded\ Word) \sum_{word\ in\ corpus} P(word) \tag{6}$$

### 3.1.3 Frequencies of word sequences:

in this approach, for performing the next words in the game, the history of the words is essential and depended on or the previous words are essential to predict the next word. In the history of the words if merely one word was utilized when the word was typed then it is called a Bigram model. On the other hand, for suggesting the next word game, if two words at the last were utilized then it is known as the trigram model (Gendron, 2015). Besides, to suggest the next word, if more than two words were utilized then it is named as the N-gram model, *N* in probability sequence is the number of the words were utilized.

It is a powerful method to conduct the next word prediction game due to it does not depend on grammar rules, therefore it is beneficial to save time.

N-gram Model is significant to make several applications, for instance, auto-complete sentences as it can be seen from Gmail nowadays. Also, it is utilized to correct spell mistakes so it is essential to check spells and correct mistakes spells. Furthermore, it can be used to check grammar mistakes in a given sentence, etc.

Frequency or probability of word sequence can be expressed step by step as follows:

a) UniGram model:

$$P(W_i | W_0 \ldots W_{i-1}) \approx P(W_i) \tag{7}$$

Bigram model, from the context, adds only a word:

$$P(W_i | W_0 \ldots W_{i-1}) \approx P(W_i | W_{i-1}) \tag{8}$$

Trigram model, from the context, adds two words:

$$P(W_i | W_0 \ldots W_{i-1}) \approx P(W_i | W_{i-2}\ W_{i-1}) \tag{9}$$

And so on for the Fourgram and Fivegram, etc.

b) Ngram Model:

$$P(W_i | W_0 \ldots W_{i-1}) \approx P(W_i | W_{i-(n-1)} \ldots W_{i-1}) \tag{10}$$

The demonstration of Ngram language model is expressed in Figure 3:

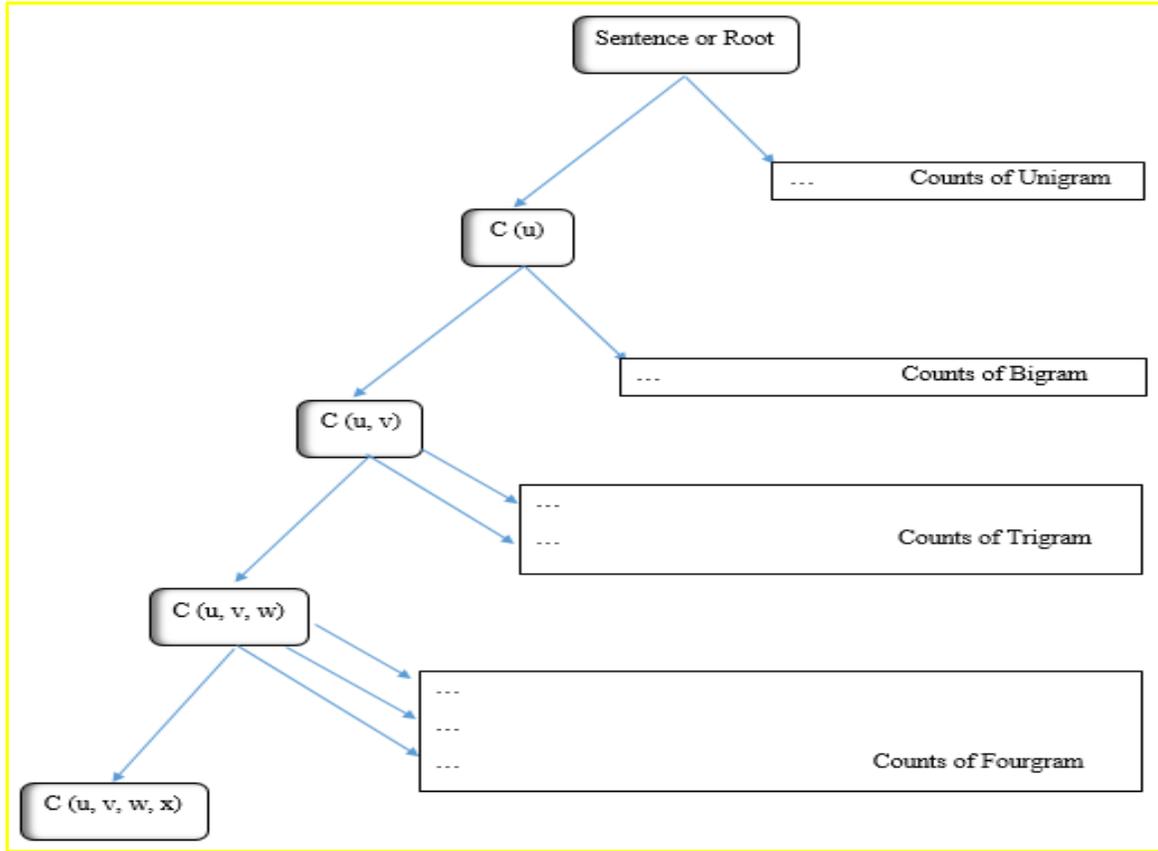

**Figure 3: illustrates the Ngram language model.**

To explain more let's take the following examples in Table 1:

Table 1: illustration of word sequence frequencies for unigram, bigram, trigram, fourgram.

| Unigram | I | Have | To | Do | My | homework |
|---|---|---|---|---|---|---|
| Bigram | I | Have | To | Do | My | homework |
|  |  | Have | To | Do | My |  |
| Trigram | I | Have | To | Do | My | homework |
|  |  | Have | To | Do |  |  |
|  |  |  | To | Do | My |  |
| Four-gram | I | Have | To | Do |  |  |
|  |  | Have | To | Do | My |  |
|  |  |  | To | Do | My | homework |

Another example is shown below in Table 2:

Table 2: illustration of word sequence second example.

| Next word prediction techniques survey paper | | | | | |
|---|---|---|---|---|---|
| **Unigram** | Next | Word | Prediction | Techniques | Survey | paper |
| **Bigram** | Next word | Word prediction | Prediction techniques | Techniques survey | Survey paper | |
| **Trigram** | Next word prediction | Word prediction techniques | Prediction techniques survey | Techniques survey paper | | |
| **Fourgram** | Next word prediction techniques | Word prediction techniques survey | Prediction techniques survey paper | | | |

### 3.2 Deep learning

Deep learning is a part of artificial intelligence that focuses on algorithms that work like the human brain to process data or it is a network that can learn from data also it is commonly named deep neural network (Dargan et al., 2019). The following subsections illustrate some of the deep learning algorithms:

**3.2.1 Next word prediction using RNN in entertainment:** people do not think on their own for every tick of time on new ideas. In the time of going through a text, each word will be acknowledged by human beings depending on their thoughts of the earlier words. They can relate and their thoughts are continuous (Bengio et al., 1994). If a classification of event types happening at each point in a novel then it is ambiguous for a neural network to utilize its analysis about earlier events to inform the next event (Rashid et al., 2019). So, the recurrent neural network can be utilized to control this issue (Mikolov et al., 2010). This means that RNN has loops that permit information to go through steps which means from one step to the next step. In another word, to conduct the next task merely the recent or current information is needed. For example, if a language model is examined attempting to suggest the next word depending on the earlier word (Huang, 2006). For instance, the last word prediction of the sentence "the sun rises in the east", other conditions will not be required due to it can be seen that east is the explicit next word in the given sentence. So, RNN has the capability of learning to utilize the previous information when the space between appropriate information and the position that is necessary is tiny (Rashid et al., 2019). On the other hand, in some cases, more condition is required. If the prediction for the last word in the sentence "I am good at playing martial arts and recently I represented my state in martial arts" is taken into consideration. So, from the last information, it is been suggested that the next word might be a sport name. In contrast, if it is desired to be known which sport is this, so, information on martial arts is needed. In this case, the space between appropriate information and the position that is necessary is large. Besides, this case is difficult to be connected by RNN (Bahdanau et al., 2015). So, in the RNN Model, the input is not examined merely, but it provides a way to examine the one earlier step. This means that the decision (*t-1*) time step has been reached straightforward and it has an impact on the next on step (*t*). In figure 4 the function of RNN or how the models work is shown:

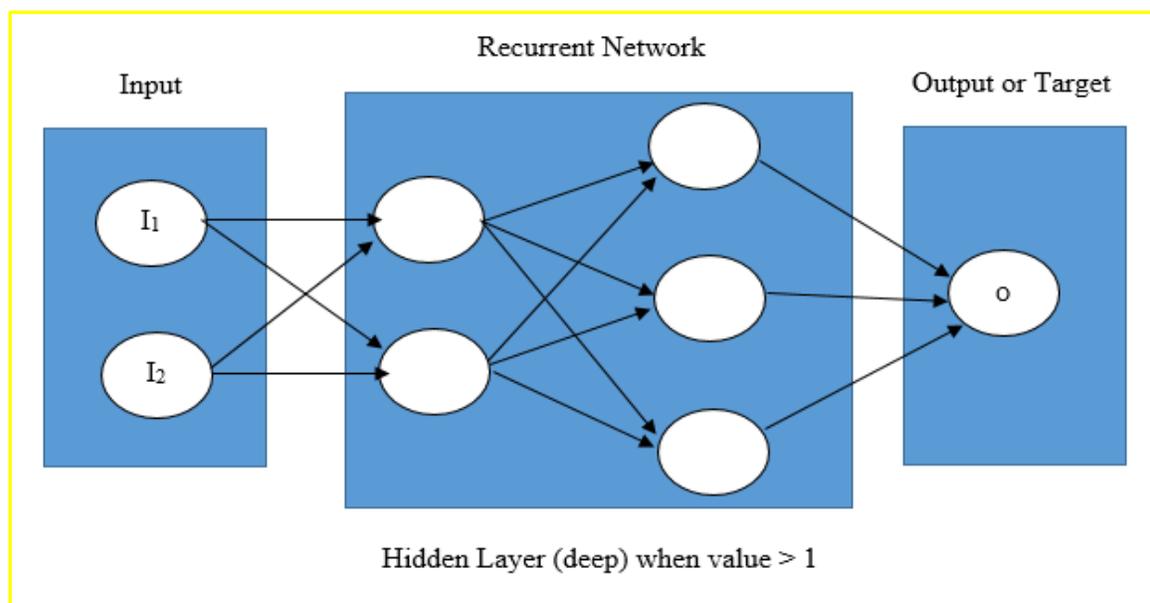

Figure 4: RNN work diagram

It is a problem to throw out whatever has been seen lastly from memory and start from the beginning each time. Let's take RNN the recurrent relation with time series or steps as it can be seen from the following equation:

$$S_t = f(S_{t-1} * W_{rec} + I_t * W_x) \tag{11}$$

Where $S_t$ represents the state at time step t, and the input is represented by $I_t$ at time *t*. Also, the weights parameters are expressed by $W_{rec}$ and $W_x$. So, the memory has been given into the model by the feedback loops due to between times steps, the information is capable to be remembered by the model. Consequently, $S_t$ which is the current state can be calculated from the current input $I_t$ and earlier steps $S_{t-1}$ by RNN. In another way, from the current input, $I_t$ and state $S_t$, the next state which is $S_{t+1}$ can be predicted. For instance, if we put 20 letters as an input of character sequence and request the model to anticipate the next character. Then, the new character will be appended and the first character will be dropped then predict again. This function will be conducted until the entire word will be predicted.

**3.2.3 Memory-based learner for next word prediction:** in this technique, a text corpus is vitally important and plays a large role in predicting the next words in entertainment (Banko and Brill, 2001). By utilizing the same text corpus in the training and testing, the next word suggestion performance will be enhanced to a great extent. In contrast, while testing if a different corpus is utilized then the performance will be reduced (Trnka, 2008). Furthermore, if the number of classes is increased which means each word is a class then the training of the machine learning algorithm is hard to conduct (Bosch, 2006). So, the most popular machine learning algorithm to be used is in this technique is the N-gram model. This deals with the occurrence probability of the $N^{th}$ given words of the last *N-1* words occurrence (Shannon, 2013). When the size of *N* is increased then the performance will be increased, however, it will be increased exponentially because of the calculation of several parameters. So, in RNN as previously stated, two considerable difficulties appear which are exploding and vanishing gradients. The RNN, traditionally a huge number of times can be multiplied by the weight matrix in the gradient signal. This means each time in the training phase the gradient signal will be smaller if the matrix weights are small, because of this reason the learning in the model will be too slow and it is known as vanishing gradient. Contradicting, exploding gradient related toa large number of weights in the matrix. This huge number will lead to learnability to have diverged. Long Short Term Memory LSTM is a method that learns from long term dependencies. The memory cell which is a new component is brought in. It consists of four parts: input gate, output gate, forget gate, and neuron. As shown in the following Figure 5:

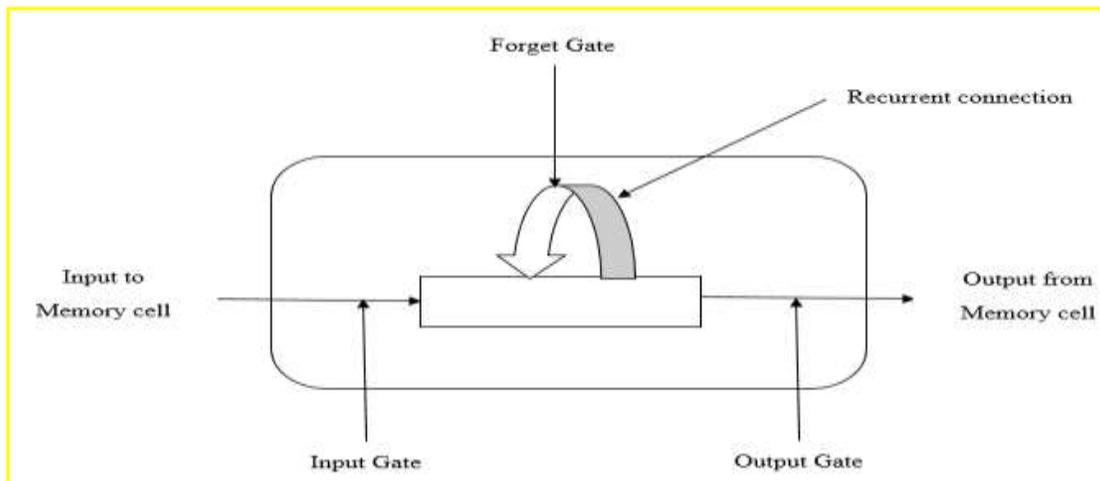

**Figure 5: the four component of a memory cell in LSTM**

So the difficulty of vanishing gradient is confronted by LSTM. This is conducted by conserving the error which is backpropagated via time and layer. Learning long term dependencies will be allowed by keeping up or continuing more constant error. In contrast, the exploding gradient will be confronted by using gradient clipping. The values of gradient clipping are not permitted to go above its predefined value. In figure 6, LSTM unit is shown:

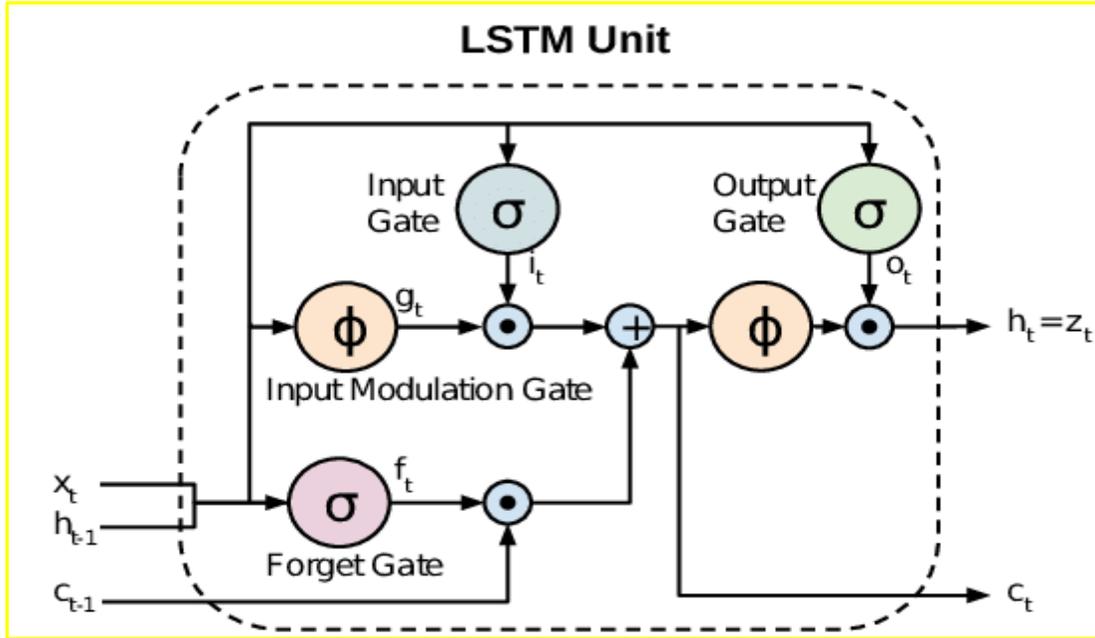

**Figure 6: LSTM unit source: (Aliaa Rassem, Mohammed El-Beltagy, and Mohamed Saleh, 2017).**

However, LSTM suffers from a large number of parameters but it resolves the problem of memory. The equations of gates of LSTM model is expressed below:

$$i_t = \sigma(w_i [h_{t-1}, x_t] + b_i \tag{12}$$

$$f_t = \sigma(w_f [h_{t-1}, x_t] + b_f \tag{13}$$

$$o_t = \sigma(w_o [h_{t-1}, x_t] + b_o \tag{14}$$

$i_t$ is an input gate, $f_t$ represents forget gate, n$o_t$ expressed as an output gate, σ is a sigmoid function, $w_x$ represents the consistent of gate x neurons, $h_{t-1}$ represents the output of earlier LSTM unit at time *t-1*, $x_t$ represents the input at the current time, and $b_x$ represents biases for the corresponding gate *x*.

The following equation shows the equation of LSTM cell state, candidate, and final output.

$$\check{c}_t = tanh(w_c [h_{t-1}, x_t] + b_c \tag{15}$$

$$c_t = f_t * c_{t-1} + i_t * \check{c}_t \tag{16}$$

$$h_t = o_t * \tanh(c^t) \tag{17}$$

From the above equations $c_t$ represents memory at a time ($t$), $č_t$ is the memory candidate at time $t$.

From the equation 15, 16, 17 it can be seen that at any time $t$, the memory or cell state knows what should be forgotten from the earlier state which means $= ft * ct-1$, and from the current time t what should be taken into consideration namely, $i_t * č_t$. Finally, the memory or cell state can be filtered by sending it via activation function. This leads to showing what should be suggested for prediction as to the output at time step t in the current LSTM unit. So, the $ht$ output can be passed to the current LSTM unit via softmax function to obtain the anticipated output $y_t$.

**3.3** The hybrid technique is about integrating multiple techniques, for instance, (Goulart et al., 2018) proposed a technique that is a hybrid model for predicting the next word, which can be used for the gaming purpose. This method depends on the Naïve Bayes model and latent semantic analysis to consider neighbor word vacant gaps. Their proposed model reached 44.2% of accuracy in completing sentences. The probabilistic method Naïve Bayes which is utilized in NLP like Ngram. It is used broadly in the machine learning era (Russell and Norvig, 2009). In this method, the joint probability is not necessarily to be utilized and build because in this technique the variables are detached into causes and effects. Though, effects are considered conditionally independent. Thus, the cost of the computation of the method will be declined. The method equation is represented below:

$$P(e|c_1, \ldots, c_n) = \frac{P(e) \sum_{i=1}^{n} P(c_i|e)}{y} \qquad (18)$$

$$\text{And } y = \sum_{j=1}^{k} P(e_j | c_1, \ldots, c_n) \qquad (19)$$

$e$ represents effects in the method and c represents causes, $y$ is a normalization factor, and $k$ represents the number of words that learned in the training model. Latent semantic analysis LSA is used to analyze the content of the texts. The LSA in the training set can be utilized to analyze the whole document or a paragraph or a phrase(Zupanc and Bosnić, 2017 ). There is a relation between words in the text corpus, so they have to be stored in a table which contains the frequency of it ($f_{i,j}$) $i$ represents any word that came into view in the level $j$. Thus, this table is necessary to compare words which can be utilized to be conducted to find similar words. So, it is essential to compute the distance between words. Therefore, LSA is utilized to calculate and infer words in a vector space. The major difficulty in LSA is the dimensions of the table due to its causes to occupy unnecessary spaces inside the memory. The following figures show the results of Naïve Bayes and LSA model:

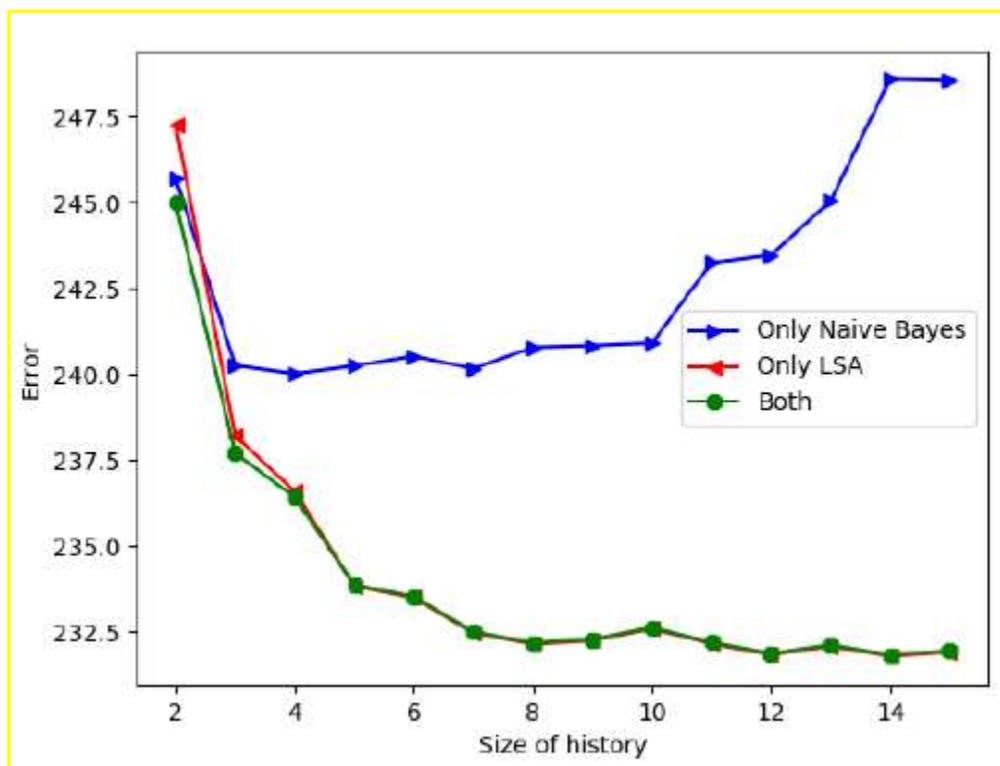

**Figure 7: represents the results of Error variation based on Naïve Bayes, LSA, and both for the size of history source: (Henrique X. Goulart et al, 2018).**

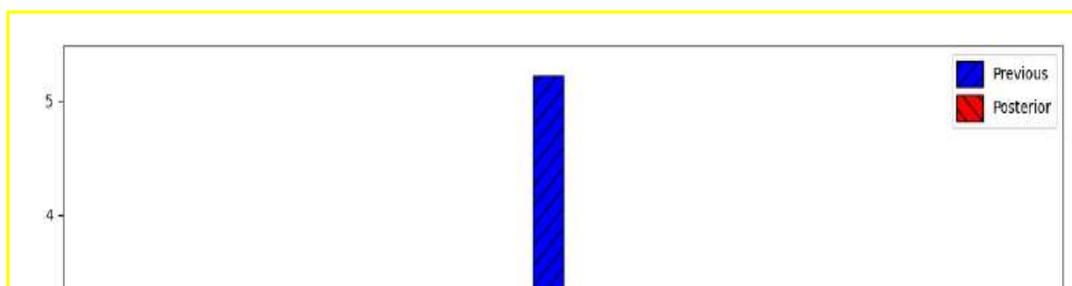

Figure 8: represents the value of lambda for the 15 gram earlier source: (Henrique X. Goulart et al, 2018).

## 4. Entertainment computing and text prediction

Entertainment computing with the next word prediction can be used as a game. Entertainment computing consists of various features, such as making games for computer, mobile entertainment, digital entertainment media, robotics entertainment, and virtual reality entertainment, etc. (Nakatsu and Hoshino, 2013). In today's world, an interesting era is entertainment computing (Wong, 2008). Recently, it was obvious that there is a concentration on entertainment and its value (Rauterberg, 2009). Besides, the next word prediction with entertainment computing could be used to make games relies on artificial intelligence to design the game (Treanor et al, 2015). Besides, entertainment computing methodology uses artificial intelligence to build games and its interface as well as using it in the background of the game (Andersen et al, 2018). Moreover, producing entertainment games using a predictive text system, which utilizes AI agents and Natural Language Processing NLP (Yannakakis and Hallam, 2006).

### 4.1 Prediction in entertainment computing in social media

Social media is one of the fields of entertainment computing. It is a type of online communication where people can compose messages, make posts, share posts, etc. for instance, Facebook, Twitter, Digg, etc. Due to using social media is simple and easy, so, public communication is changing fast in society, environment, politics, and in the entertainment with technology. All of these because of using social media.

Because social media could also be deduced in the form of common sense collection, a decision was made to consider its value at real-world prediction consequences. It is essential to predict in social media because, firstly, if we compare prediction in machine learning with human work, the cost with prediction in machines is lower (Bothos et.al, 2010). Secondly, the probability of words with low and high rates is not well predicted by people due to people contribute to miscalculate high probabilities with undervalue and low probabilities with overvaluing (Wolfers and Zitzewitz, 2004). Thirdly, people make a decision either intentionally or unintentionally affected by their interest, profit, and ambition, not based on accurate probability (Wolfers and Zitzewitz, 2004). Lastly, the prediction approaches could handle the massive amount of data then respond provided fast (Hanson, 2004).

Many subjects can be applied with social media but the prediction with social media has to meet the following requirement:

In the first, the subject with prediction must relate to human events. Users, share their thoughts and opinions on social media. So, the information will be analyzed, extracted, and integrated by prediction approaches. After that, the prediction should be made based on the influence of people on the predicted subjects.

Secondly, in the case of involving a huge number of people, the distribution composition on social media for involved people has to be at least similar to the real world (Huff, 1993). This is because not all people around the world use social media, so the users who use social media can be considered as a sample. However, the process of samples is not handable, therefore, it may cause of bias with sampling. Thus, bias cannot be excluded from sampling entirely. Instead, ensure that the relative amount of biased samples is in agreeable range.

Finally, the texts that involved in social media have to be simple, which are composed publicly, alternatively, the social media contents will produce bias (Huff, 1993). If an example is taken into consideration, there is social consensus, which gives suitable and good tips, contradicts, extravagantly low tipping is unarguable. For the huge number of people on social media, it is not acceptable to admit the low paid tips. The response could be obtained by using anonymous mode but this mode has no relevant information on social media structure.

### 4.1.1 Social media message characteristics predictor

Characteristics of messages concentrate on the messages, for instance, time series and sentiment measurement. In case all objects are concentrated on then all available posts, with timestamps fetched. Contradicts, the man-crafted keywords in consequences of the search will be preferred, all these will be beneficial for prediction, this depends on texts or posts based on that the most frequent words can be predicted what word is most likely to happen then the next word.

**4.1.1.1 Sentiment measurement:** sentiment measurement is a fixed part of the posts. Besides the word prediction might be for a specific sentiment or category, for example, diseases, happiness, places, etc. due to they are not investigated thoroughly. Thus, the idea is to extract and utilize them as a general one.

In the sentiment measurement analysis with quality, the texts can be predicted for positive, negative, or neutral, however, they have to be labeled. As a consequence, we have five predictor factors; negative, positive, non-neutral, neutral, and total. These are the five agreeable predictors.

These predictor factors can have its value in different places, for instance, predicting the next best movie, the number of positive votes equates with the event consequences is better than counting total in the pre-event time. In contrast, in the post-event time, it is better to conduct counting total (Mishne and Glance, 2006). Also, the ratios can be calculated among them. It includes the ratio between the accounting positive and posts in total (Zhang and Skiena,

Cite as : Hamarashid, H.K., Saeed, S.A. & Rashid, T.A. A comprehensive review and evaluation on text predictive and entertainment systems. *Soft Comput* (2022). https://doi.org/10.1007/s00500-021-06691-42009). Also, it contains the calculation of the ratio between non-neutral and neutral posts, also, it includes the computation of ratio between positive and negative posts (Asur and Huberman, 2010). The mentions ratios reverse the correlated strength of the mentioned sentiment measurement. The sentiment difference and sentiment index can be calculated by combining the basic aspects (Zhang and Skiena, 2009). The following equations represent the calculation of sentiment difference and index:

$$setiment\ dif = \frac{N_{positive} - N_{negative}}{N_{total}} \tag{20}$$

$$sentiment\ Index = 100 * (\frac{\frac{N_{positive} - N_{negative}}{N_{total}}}{2} + 0.5) \tag{21}$$

$N_{positive}$, $N_{negative}$, and $N_{total}$, represent positive, negative, and total posts consecutively. On the other hand, the sentiment index was strongly recommended to and useful to be used for predicting the name of Oscar winner when it is used for box-office movies due to it was proved.

Overall, the text or an event can be predicted either positive or negative, it can be changed based on the owner desire, for example, if we have the words "I" "am" the predicted word will be "Happy" then this indicates that the post is positive.

**4.2 Prediction in entertainment computing (games)**

Next word prediction can be utilized as an entertainment game. Text prediction interfaces that auditor text while it is typed by a user then it is providing a set of anticipated next words for the user to choose straightly. This is can be conducted on smartphones. A small algorithm interference can be represented in the action of text composition that allows smartphone users to interact with the system.

Artificial intelligence game-based design is an approach to designing a game depending on artificial intelligence agents. This can be used for the interface of the gameplay and the background. Thus, it is helpful for players' attention due to it is a memorable part of the roles and management of AI methods (Treanor et al, 2015). Digital games particularly are well known to analyze artificial intelligence algorithms due to it consists of codes, so they could embed playable form or version of the approach directly. So they are engaged to conduct comments on them, therefore, a chance will be given to the player to experience the interaction of the game to see what it looks like through these algorithms. Besides, a system, for example, Say Anything (Swanson and Gordon, 2008) has favorably conducted artificial intelligent methods to build playable cooperative experiences in writing games. Possibly, artificial intelligent game-based design approaches might be practiced with artificial intelligence algorithms, which act a common role in the player during every day playing lives. The prediction texts, which are used in a game mechanic as an example, might appear impressive questions algorithmic game into the mind of the players. Following is the representation of such a game.

Predictive texts as a game mechanic taken into consideration as an example of the entertainment game, which uses text predictive system, a player can interact with the game by composing a message or inputting a word as a target word in the game. Thus, to do progress, the player asked to enter the anticipated text in the game. In this step, the player will be expedited by the system by providing a predictive text to the interface of the game. This helps the player to look at what has been composed recently. Also, in the game real-time feedback was provided to select the words that most likely to utilize in the next step (Campton et al, 2015). The player might choose one of the suggested words at any time during writing a message. The predicted words to the user rely on word frequencies that are drawn out from the text corpus of the entire tweets that exist. The game utilizes the Markov Chain model to train the data set of the corpus of the entire tweets to assist the player in the text in the game style.

The game content formed as a partly arranged sequence of broadly contained in itself vignettes. Every vignette is formed in the style of dream vision. This predicts several events or scenes for the future. Whenever the player tries to input the social media post into the game by utilizing the next word prediction interface system, one of the predicted words might be assigned to a vignette randomly, which is suitable for the player progress in the game. Thus, choosing the predicted words frequently will lead the player to accomplish vignette text, each word separately, into the textbox or text input. Due to each predicted words could be chosen from vignette or the corpus that presented in tweets and also due to that the player has unintended control on which words will be predicted, thus, the game design provides ambiguity to the corpus, vignette or the player for the last content of the messages that they recently composed.

One of the expected consequences of the next word prediction game design is to increase the players understanding that they will gain wonderful encouragement whenever they depend on the next word predictive game to input texts. This will be stimulated in which the next word prediction interfaces are consistently utilized to conduct artificial intelligence-based design, for example, in the form of a text prediction game, which is mostly playable on social media. In this game, the player or participant attempts to input or write the beginning of a sentence manually, so the next word prediction will be utilized to complete the rest of the sentence and then share the consequences (Neal, 2017).

By inspiring the players to engage with the next word prediction games and by isolating the players from common features for writing in next word predictive system interface, the main goal of this game was to get into consideration to the way of that the player input text, how it is affected by algorithmic next word prediction system. The players can be a success in the game and they can leave and return to interact with the next word predictive game whenever desired.

## 4.3 Utilizing NLP technique for search recommendation in entertainment computing

The idea of this technique is to mix the search and recommendation of entertainment computing. This will be helpful for users to utilize natural language processing to conduct different tasks on digital media content and information retrieval. Recently, mobile and tablets are developed quickly to derive entertainment computing, such as digital media. Simultaneously, robust technology exists for speech recognition to convert voice to text. This enhancement enables voice interfaces and NLP to be usable for a variety of tasks, for instance, inputting text to turn on the electric devices or using

TV remote for playing games, such as word prediction games with the TV features. So, the NLP interface can be used to conduct different functionalities in entertainment computing like video and music, which related to the user tasks that connected with digital media. For instance, retrieving search consequences and recommendations, handling common TV commands, retrieving information from the digital media content, and predicting answers for trivial questions that are related to.

This is about making questions and answers system. The goal of this system is to provide accurate answer texts to the queries of particular NLP users instead of classical search consequences, which give corresponding documents. Many of these systems mostly rely on ontology, which has the knowledge-based data that shaped and arranged by an ontology. The users in the system can input questions and then the system will give precise answers to the user when the question is analyzed then the answer is extracted (Athira et al, 2013). Knowledge-based ontologically, dedicate an appropriate approach to consolidate semantic users considered demanding, but at this point, NLP must be organized to the form of ontology statements, for example, Siri by Apple. In the entire of these types of systems, the queries of NLP are first translated to the form that is appropriate with ontology then this form will be utilized to discover the final consequences. The structure of the system is shown in Figure 9.

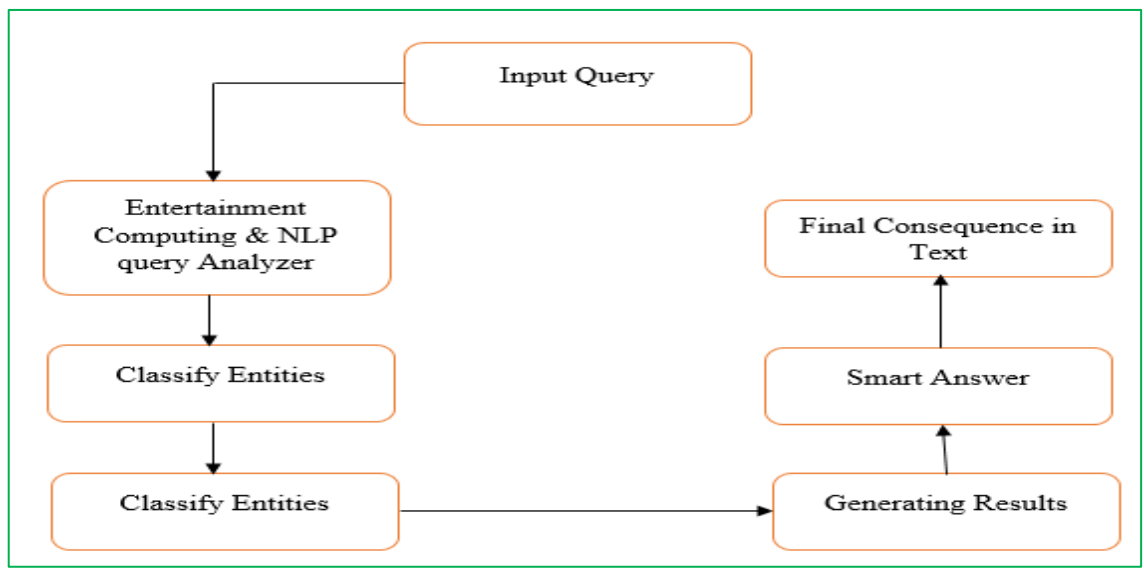

**Figure 9: represents the system structure and steps of NLP entertainment computing search and recommendation**

## 5. Analysis of the next word prediction techniques

The next word prediction system is supported in the typing system. If the efficiency is taken into consideration, then it will be seen that the main goal is to reduce effort and message typing time. So, if it assists people in reducing the necessary efforts to type, it is essential to decline the keystroke number for writing a text message. If it is helpful for people in declining the number of letters that made on a time, several integrated letters in a text. In other words, the number of letters, which is written by only one selection should be smaller than the individual prediction (Garay and Abascal, 2006).

If a simple example is taken to illustrate the result of a saved keystroke in percentage for next word prediction in the English language, as shown in Table 1:

Table 1: the percentage of saved keystroke results for next word prediction in the English language

| Grams | Results |
|---|---|
| Bigram and trigram | %28.2 |
| 2, 3, 4, 5, 6 grams | %37.4 |
| Plus personal lexicon | %47.7 |

The first result shows that it depends on the agreement of the texts from test and training with the availability of a chunk of information sources.

The second result illustrates that it relies on the agreement of the texts used in the test and texts used in the training for the n-gram. This consistently enhances the consequences.

The last result reveals the power of a personal lexicon when it is utilized.

### 5.1 Prediction results

This part reveals the discovered appropriate results in the literature (Vitoria and Abascal, 2006).

The consequences that obtained from the word suggestion system it was represented in (Sharma and Samanta, 2014). It is noticed that the projected software makes %95.6 corrections in simulation errors. In general, on average it reaches % 62.50 in saving keystrokes with a hit rate of %96.50. The produced system can be utilized easily. It does not need the experience to use the system.

The statistical word suggestion system as an example of Lipik which is used for composing text messages. In the system, a virtual keyboard was made to input text messages. When a user types a word then a list of next word suggestions will be shown after selecting the word by pressing the space button then the next possible words will be pop up (Vitoria, Abascal, 2006). The calculation of input text rate in word per minute WPM is 5.0 for users with no experience and 5.3 with experienced users.

Google made a prediction system to suggest the next words while searching and shows the consequences of the prediction in the suggestion words window. When a user inputs a part of a word then the system gives the next words most likely to suggest in the prediction window. Multi words suggestion is also available in the prediction system and it can be seen in the suggestion window. It can explore from the internet and fetch the consequences (Vitoria, Abascal, 2006). The text rate input in WPM by Google for the Hindi language is reached 5.3 with no experienced users and 5.6 with experienced users.

Google transliterate and Quillpad is bilingual sound predictive software. This system can be utilized to trans-literately predict between two languages, which are based on alphabets after the system is trained (Prakash, 2012). The system shows the prediction when a user types. The prediction words are ordered based on the declining order depending on their statistical probabilities and frequencies.

Table 2: the word suggestion systems comparison

| Unit | Google | Lipik |
|---|---|---|
| Saved keystrokes | %16.8 | %32.4 |
| Usage of prediction | %87.1 | %85.7 |
| Hit rate | %28.5 | %85.7 |
| Word input rate (word per minute) | 4.8 | 7.3 |

Several systems were designed and implemented to predict the next word one of those systems is produced in the Urdu language with the evaluation of word prediction which conducted on 20 students. The quickness of inputting text on average was 13.4 words per minute. The highest rate of text entry was 22.5 words per minute.

In the following table, the word prediction accuracy is shown:

Table 3: next word suggestion accuracy

| Kinds of predictions | 1 to 3 grams | 3 to 6 grams | 6 to 9 grams |
|---|---|---|---|
| Half input word | %46 | %65 | %68 |
| Next word | %52 | %70 | %71 |
| **Prediction with modeling** | | | |
| Half input word | %52 | %75 | %82 |
| Next word | %68 | %80 | %85 |

Word prediction systems were conducted to increase the speed of typing and to decrease the efforts to type a word while composing a text message. However, disabled people get benefit from these kinds of systems, people with no disabilities can utilize these types of systems to help them input texts correctly due to the system is also helpful with correcting spelling mistakes.

For the **RNN** two different data sets were produced. Firstly, the data set includes 1037 words which are taken from the Bhanumati novel and the words were increased by duplicating the words from the novel five times inside the dataset. Secondly, the first data set phonetic transcription is included in the second data set word by word until it is reached the end and it consists of 1076 words. In the RNN approach, the two data sets were fed in and the consequences were computed after every 1000 epochs. So, for very1000 epochs, the consequences were too long, therefore, only the epochs which are getting the highest rate of accuracy are shown according to the computation. During the testing RNN model, merely three random words are fed into the RNN approach in sequence and suggesting the next word immediately. In Table 4, the test of the technique with the highest rate of accuracy with its epochs is represented:

Table 4: a tested model with the highest rate of accuracy with its epoch

| Number of tests | Hidden layer | Neurons inside every layer | Learning rate | Epoch | Accuracy |
|---|---|---|---|---|---|
| Test 1 | 2 | 128 | 0.001 | 98000 | %88.2 |
| Test 2 | 2 | 256 | 0.001 | 82000 | %77.2 |
| Test 3 | 2 | 512 | 0.001 | 86000 | %70.4 |
| Test 4 | 3 | 128 | 0.001 | 100000 | %81.1 |

| Test 5 | 3 | 256 | 0.001 | 58000 | %74.3 |
| Test 6 | 3 | 512 | 0.001 | 50000 | %71.9 |
| Test 7 | 4 | 512 | 0.001 | 98000 | %70.3 |
| Test 8 | 4 | 512 | 0.003 | 74000 | %22.2 |

The first data set is fed into various models without transcription to produce a different structure of 100000 epochs. According to Table 4, the highest rate of accuracy was % 88.2 with the 98000 epochs for the first data set when it has two hidden layers and every layer has 128 neurons with a learning rate of 0.001.

Correspondingly, for the second data set which includes a phonetical transcription of the words was fed into a variety of models. Each model was run for 100000 epoch and it obtained the highest rate of accuracy for each approach as it is illustrated in Table 5.

Table 5: represents the second dataset tests for the highest rate of accuracy with its epoch

| Number of tests | Hidden layer | Neurons inside every layer | Learning rate | Epoch | Accuracy |
|---|---|---|---|---|---|
| Test 9 | 2 | 128 | 0.001 | 100000 | %46.1 |
| Test 10 | 2 | 256 | 0.001 | 100000 | %67.0 |
| Test 11 | 2 | 512 | 0.001 | 77000 | %50.6 |
| Test 12 | 3 | 128 | 0.001 | 68000 | %56.4 |
| Test 13 | 3 | 256 | 0.001 | 77000 | %72.1 |
| Test 14 | 3 | 512 | 0.001 | 98000 | %63.6 |
| Test 15 | 4 | 512 | 0.001 | 91000 | %70.4 |
| Test 16 | 4 | 512 | 0.003 | 34000 | %16.1 |

From Table 5, merely the highest rate of accuracy with its epoch is shown because the consequences for each epoch with 1000 will be huge. Thus, only the epochs with the highest rate of accuracies are represented in Table 5. Consequently, the highest rate of accuracy was obtained with %72.1 with 77000 epochs with three hidden layers and every layer has 256 neurons with a learning rate of 0.001.

Overall, from the earlier two tests, it can be seen that the test without the transcription in Table 4 the accuracy on average is declining when the number of neurons is increased in every hidden layer in the case when the number of hidden layers is fixed. Also, the learning rate and the number of hidden layers affect the accuracy of the average. By increment, the number of the hidden layer the accuracy was declined. Moreover, when the number of learning rates was raised from 0.001 to 0.003 it can be seen that the accuracy is declining drastically. In contrast, the test with the transcript texts it can be noticed that the accuracy goes up when the number of neurons is growth from 128 to 256 for every layer. On the other hand, when the number of neurons is incremented to 512 then the accuracy is declined. Indeed, when the number of the hidden layer is changed from 2 to 3 then the accuracy goes up but attempting to increase hidden layers furthermore decrements accuracy. Likewise, the accuracy is drastically reduced when the number of learning rate is changed from 0.001 to 0.003.

The purpose of using memory-based learners for next word prediction is being able to utilize a larger context which means a larger *N* number in N-grams. In this method, if a context could not be found or merely found in a few times this will not lead to predicting the next word in this case back off algorithm needed to be utilized. This will be conducted by receding to lower *N*. The correct prediction will be made when all the preceded words have been saved in order.

So, testing this algorithm with bases on the N-gram model then the model receding to a lower number of *N* in case, not adequate texts are encountered. Then, it was seen that the performance of the system did not alter in the case that *N* is more than 3 or trigram this is happened due to the size of the training set was small.

For the accuracy of the prediction different size was tested the best prediction accuracy rate is obtained where *N* is equal to 3 or 4.

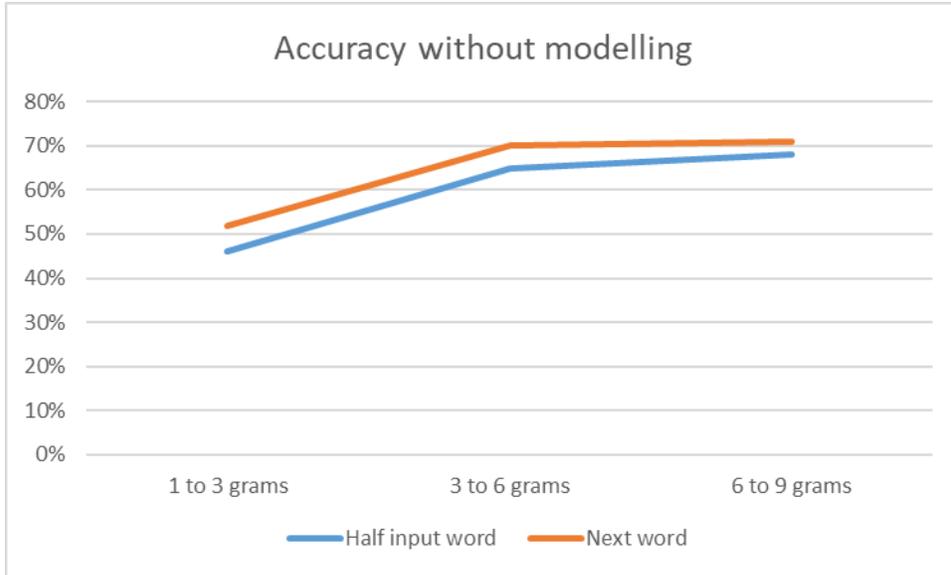

Figure 10: Accuracy without modeling

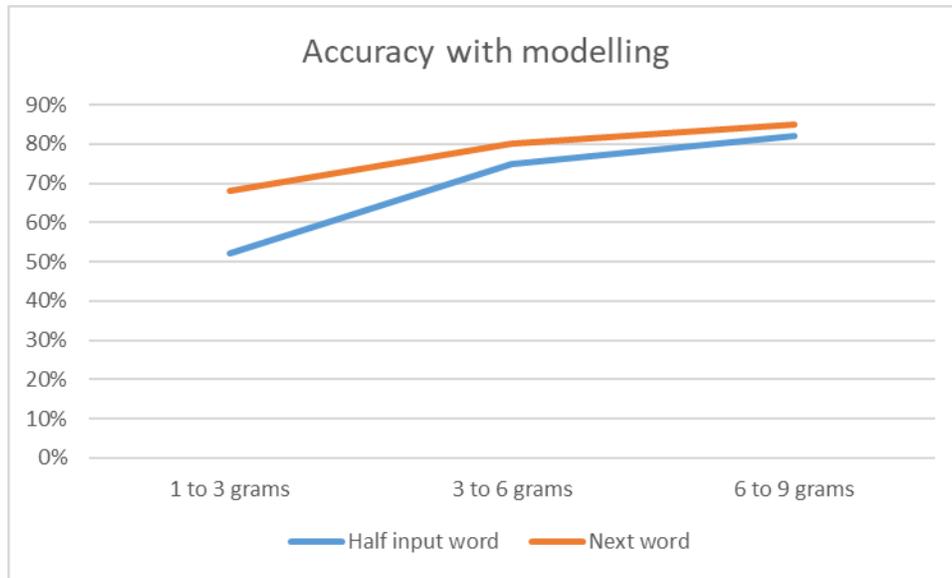

Figure 11: Accuracy with modeling

NLP with entertainment computing performance and accuracy test, the user queries were tagged with anticipated entities, intent, and predicted consequences into the practical system. In this system, a random sample was collected, which is nearly 22K queries from different users. Table 6 explains the ratio of the accuracy of the system, which is conducted with the test set in the question and answer system across different phases. In the test part, the ratio for intent recognition is above 90 percentage and the ratio for entity recognition is more than 80 percent. The incorrect ratio is because of misspelling especially in the entity recognition in the speech to text difficulties.

Table 6: represent the accuracy percentage for the 22K quires

| Types | Succeeded Ratio |
|---|---|
| Entity recognition | 81 percentage |
| Intent recognition | 90 percentage |

## 6. Research open problems

Next word prediction is a continuously interesting and challenging problem. Every day the availability of data will increase, so new challenges will be faced according to process the provided data to be used in predicting the next word. The challenges of next word prediction include long- term predictions. According to long term prediction dependencies, most of the literature on next word prediction and analysis claiming that some of the proposed methods and algorithms can be utilized to predict the next word but not for adequate long term prediction. It is massively demanded to be conducted due to there might be algorithms work fine on predicting the next word but the main challenge is predicting for a longer-term. However, some algorithms and techniques developed to conduct long term dependency prediction but they may suffer from several points for instance LSTM. Although LSTM utilizes as a memory to predict longer due to it is using a history of words to predict for the long term but LSTM suffers from having too many parameters this causes computation cost and time-consuming. This means that the performance will be decreased.

Several techniques developed to predict the next word but each of them suffers and has its disadvantage. For example, Markov Chain, frequency of words, predicts the next word but suffers from long term prediction. On the other hand, the developed algorithms sometimes work for long term prediction but the performance problem will arise and the computation time will increase. Namely, although there were efforts to develop techniques for predicting the next word for longer-term the challenge of efficiency and accuracy will come into being.

Researchers, mostly focused on short term prediction in comparison to long term prediction, for instance, 3grams in n-gram model. Nevertheless, it is capable to predict next word but for predicting sentences it is quite useless. Though some algorithms were enhanced such as RNN and LSTM for long term prediction they need powerful computers to execute. Another point is about datasets huge datasets need more powerful computers to process. Also, if extra texts needed to be extended to the corpus then the training and testing of the model required to be conducted again. This step is essential to be executed because later the efficiency and accuracy of the model will be determined based on the training and testing the model.

Recently, because of the importance of predicting the next word with mobile phones or search on the web or with computer software even with various languages, more attention is being considered. In the first step, the next word prediction depends on a text corpus, so for a specific language a good corpus should be provided or the sources of collected texts for a specific language should be good. Also, the number of texts should be taken into consideration due to it has an impact on processing it because of this, the corpus needed to be preprocessed. On the other hand, to predict for long term dependency, although good techniques exist like LSTM to predict for a longer-term too many parameters reduce the efficiency of the model. It is better to reduce the number of parameters to enhance the performance and execution time because it has several gates. It is better to update and replace data instead of throwing the forget data.

Entertainment computing with the prediction barriers could be using simple linear approaches, which is suitable merely for some conditions or specific conditions. On the other hand, social media is complicated when it is divided or distinguished by parameters, therefore, the predictors and prediction consequences on social media are a non-linear correlation. Besides, the mixing of the approaches could lead to enhancement. This mixing approach, such as a trained neural network instantaneously, enhanced to a new approach on social media. Besides, deep learning concentrates more on long terms. Also, non-linear approaches should be tried to discover a reasonable approach for every prediction domain. Another point is that there are various types of prediction items, which represent various parts. If the recommendation is taken into consideration, the recommendation for DVDs is more acceptable than books. On the other hand, there is still no entirely agreeable consequence on why the differences among them happen. This problem exceeds barriers for modeling. So, modeling formally could be essential and significant to comprehend and examine the parts and attitude of prediction approaches.

## 7. Conclusion

There are several approaches to the next word prediction that intermittently designed. The main goal is to speed up typing, reduce efforts, and consuming the time for composing a text message also to boost the communication rate. These approaches mainly produced to help people with disabilities or dyslexia also people with no disabilities can utilize next word prediction systems to help them to correct spelling mistakes and type their desired words with few efforts during composing text messages in other words it needs less typing words.

Various techniques have been proposed to enhance the level of next word prediction systems. This paper has addressed the analysis of these approaches. Also, various computations of next word prediction models have been represented especially perspective of view in saving keystrokes, performance, hit rate, input text rate, and accuracy.

As a consequence, depending on the results of the discussed next word prediction approaches, the N-gram model is better to utilize because of the following reasons; firstly, it is straight forward and easier to apply compared to the other techniques due to it depends on text corpus merely. Secondly, by utilizing the N-gram model, better consequences obtained with 3 or 4 grams in other word trigram or four grams. However, few data leads to predict less some times. So, in this case, the bakeoff algorithm is used due to whenever inadequate prediction happens then it will go a step back from *N* to *N-1* for instance from fourth gram to trigram. Thirdly, better accuracy was obtained, nevertheless, by increasing the number of N-grams the performance slightly declined because of the computation time. Lastly, the N-gram model only needs to create or to make grams. On the other hand, the other techniques required to train and test the data, whenever more data is needed to be included then the data again needs to be trained and tested. This requires more time and reduces performance and the accuracy of the prediction is similar.

In the future, for selecting a better system for next word prediction, so usability tests model may be utilized to determine suitable next word prediction system.